\documentclass[twoside]{article}

\usepackage[utf8]{inputenc}
\usepackage[english]{babel}
\usepackage{pdfpages}
\usepackage{xcolor}
\usepackage[font=it, labelfont =normalfont]{caption}
\usepackage{amsmath}
\usepackage{amsthm}
\usepackage{amsfonts}
\usepackage{amssymb}
\usepackage{mathrsfs}
\usepackage{mathtools}
\usepackage[titletoc]{appendix} 
\usepackage{float}
\usepackage[separate-uncertainty = true]{siunitx}
\usepackage{array}
\usepackage[version=3]{mhchem} 
\usepackage{comment}
\usepackage{enumitem}
\usepackage{booktabs, multirow} 

\usepackage{pdflscape}
\usepackage{afterpage}
\usepackage{tabularx}
\usepackage{longtable}
 \usepackage{url}
 \usepackage{multicol}
\usepackage{hyperref}
\usepackage{xcolor}
\usepackage{microtype}
\usepackage{graphicx}
\usepackage{subfigure}
\usepackage{booktabs} 
\usepackage{hyperref}

\usepackage{natbib}
\setcitestyle{authoryear, semicolon, open={(},close={)}}

\usepackage[accepted]{aistats_style}
\usepackage{subfiles}
 
\global\long\def\b#1{\boldsymbol{#1}}

\global\long\def\br#1{\left( #1 \right) }
\global\long\def\sqbr#1{\left[ #1 \right] }

\begin{document}

\twocolumn[

\aistatstitle{Efficient exploration in binary and preferential Bayesian optimization}

\aistatsauthor{ Tristan Fauvel \And Matthew Chalk}

\aistatsaddress{Sorbonne Université, INSERM, CNRS, Institut de la Vision \\
    F-75012 Paris, France} ]

\begin{abstract}
 Bayesian optimization (BO) is an effective approach to optimize expensive black-box functions, that seeks to trade-off between exploitation (selecting parameters where the maximum is likely) and exploration (selecting parameters where we are uncertain about the objective function). 
In many real-world situations, direct measurements of the objective function are not possible, and only binary measurements such as success/failure or pairwise comparisons are available. To perform efficient exploration in this setting, we show that it is important for BO algorithms to distinguish between different types of uncertainty: epistemic uncertainty, about the unknown objective function, and aleatoric uncertainty, which comes from noisy observations and cannot be reduced. In effect, only the former is important for efficient exploration. Based on this, we propose several new acquisition functions that outperform state-of-the-art heuristics in binary and preferential BO, while being fast to compute and easy to implement. We then generalize these acquisition rules to batch learning, where multiple queries are performed simultaneously. 
\end{abstract}
\section{Introduction}

Bayesian optimization (BO) is a powerful method for optimizing black-box functions: that is, functions whose analytic form is unknown, and for which no derivative information is available  (see \cite{Brochu2010a, Shahriari2016, Frazier2018} for review). Bayesian optimization is especially important when function evaluation is expensive in terms of some limiting resource such as time, money, or computational resources. To perform efficiently, BO seeks to balance exploitation, where new queries are chosen where the expected value of the function is large, and exploration, where new queries are chosen in regions of high uncertainty.

In many real-world problems, direct evaluation of the objective function is impossible, and only binary outputs such as success/failure or pairwise comparisons are available.  Examples include the tuning of a robot’s gait parameters, where the robot may succeed/fail in moving over an obstacle \citep{Tesch2013}, or the optimization of machine learning models' hyperparameters, by early-stopping the training of under-performing models while continuing others \citep{Zhang2019a}.
The generalization of BO to these scenarios is called binary Bayesian optimization (BBO) \citep{Tesch2013}. 

To perform efficiently, BO algorithms seek to explore regions of the parameter space where there is high uncertainty. In standard BO with noiseless observations there is only one source of uncertainty: epistemic uncertainty about the value of the objective function, $f\br{\b x}$, given a search parameter, $\b x$. In contrast, in BBO, observations ($c=0/1$) correspond to stochastic samples from a Bernoulli distribution, $P\br{c=1|\b x} = \Phi\br{f\br{\b x}}$, where $\Phi$ is the normal cumulative distribution function. As such, there is an additional source of uncertainty: the irreducible aleatoric uncertainty about the binary observation, $c$, given a known objective function, $f\br{\b x}$. Further, and in contrast to continuous BO with homogeneous noise \citep{Letham2019}, the magnitude of this aleatoric uncertainty depends critically on the objective function: for known $f\br{\b x}$, it is largest when $f\br{\b x}= 0$ (and thus, where $P\br{c=1|\b x}=0.5$), and smallest when $f\br{\b x}\gg 0$ or $f\br{\b x}\ll 0$ (and thus, where $P\br{c=1|\b x}\approx 1/0$).  

We show that to explore efficiently, BO algorithms should distinguish between both types of uncertainty. In effect, optimal exploration requires sampling from regions of high epistemic, but not aleatoric, uncertainty. We use this insight to design new acquisition functions for BBO and preferential BO (where observations correspond to comparisons between different parameter settings) that: (i) are efficient and easy to implement, and (ii) out-perform state-of-the-art algorithms, which often mixed up different types of uncertainty \citep{Brochu2010b}, resulting in sub-optimal exploratory behavior \citep{Nielsen2015a}. Finally, we show how this idea can be generalized to deal with batch learning, where comparisons between multiple different parameter settings are performed at once.


\section{Inference in Bayesian optimization with binary outputs}
Bayesian optimization (BO) aims to solve the following global optimization problem:
\begin{equation}
\label{problem}
\mathbf{\boldsymbol{x^\star}}=\underset{\boldsymbol{x} \in \mathcal{X}}{\arg \max } \ f(\boldsymbol{x}),
\end{equation}
where $\mathcal{X}$ is a bounded search-space. The aim is to find the global optimum, $\b x^\star$, using few function evaluations. 

The first step of BO is to learn a model of the objective function using previous observations, $\mathscr{D}$. This surrogate model describes the probabilistic belief about the objective, $p\br{f|\mathscr{D}}$, and can be used to incorporate prior knowledge. Gaussian process (GP) models are usually chosen because of their flexibility, their practicality in incorporating prior knowledge, and because they are considered to provide a reasonable estimation of uncertainty \citep{Rasmussen2006}.

\subsection{Binary Bayesian optimization}
In binary Bayesian optimization (BBO; \citet{Tesch2013}), the result of each evaluation, $c$, is binary (0/1), and determined by the objective function, $f$, as follows:
\begin{equation}
    P\br{c=1|f, \b x} = \Phi\br{f\br{\b x}},
\end{equation}
where $\Phi$ is the normal cumulative distribution function.  Given a GP prior and Bernoulli likelihood, the posterior distribution, $p\br{f|\mathscr{D}}$, 
is usually approximated using Laplace approximation or Expectation Propagation \citep{Minka2001, Seeger2002, Rasmussen2006}. Consequently, the posterior distribution $p(f|\mathscr{D})$ is approximated by a GP, with mean $\mu_f\br{\b x}$ and variance $\sigma^2_f\br{\b x}$. The probability of observing $c=1$, given $\boldsymbol{x}$ can then be expressed analytically as:
 \begin{eqnarray}
P\br{c=1|\b{x}, \mathscr{D}} &=& \mathbb{E}_{p\br{f|\mathscr{D}}}[\Phi(f(\boldsymbol{x}))]  \\
 &=& \Phi\left( \frac{\mu_f(\boldsymbol{x})}{\sqrt{1+\sigma_f^2(\boldsymbol{x})}}\right). \nonumber 
 \end{eqnarray}
We herein define $\mu_c(\b x) \equiv P(c=1|\b{x}, \mathscr{D})$.
 
\subsection{Preferential Bayesian optimization} 
In preferential BO (PBO, \citet{Brochu2010b, Gonzalez2017a,Dewancker2018}) the result of an evaluation, $c$, depends on the relative value of the objective function, $f$, with two different parameter settings, $\b x$ and $\b x'$, according to: 
\begin{equation}
    P(c=1| f, [\boldsymbol{x}, \boldsymbol{x}']) = \Phi\br{ f\br{\b x}- f\br{\b x'}}.
\end{equation}
The parameters to be compared, $\b x$ and $\b x'$, are called a duel. Preference-based optimization occurs in many applications where the value of the objective function cannot be measured directly, in particular in applications relying on human feedback, since comparison is often easier for humans than rating  \citep{Kahneman1979, Shah2014}.

PBO can be treated as a special case of GP classification \citep{Houlsby2012}, where $
    P(c=1| f, [\boldsymbol{x}, \boldsymbol{x}']) = \Phi\br{g \br{\b x, \b x'}}$ and $ g$ is  a `preference function'  defined as $g\br{\b x, \b x'} = f\br{\b x}-f\br{\b x'}$. As before, the objective function $f$ is modelled using a GP prior with mean $m\br{\cdot}$ and covariance $k\br{\cdot, \cdot}$. This results in a GP prior for  $g$, with mean and covariance:
\begin{eqnarray}
m_{g}\br{\sqbr{\b{x},\b{x}^\prime}} &=& m(\b{x}) - m\left(\b{x}^\prime\right), \\
k_g\br{\sqbr{\b x_i, \b x_i'}, \sqbr{\b x_j, \b x_j'}} &=& k\br{\b x_i, \b x_j}   + k\br{\b x_i', \b x_j'} \\
&&- k\br{\b x_i, \b x_j'}  - k\br{\b x_i', \b x_j}. \nonumber 
\end{eqnarray}

\section{Exploitation \& exploration in Bayesian optimization with binary outputs}
\label{exploitation}
Having inferred a Bayesian model of the objective function, $p\br{f|\mathscr{D}}$, the next step of any BO algorithm is to select a new parameter value, $\b x$, to evaluate the objective. This is typically done by maximizing an `acquisition function',  which implements the trade-off between exploration and exploitation. To design an acquisition function for BO with binary outputs requires a clear view of what exploitation and exploration are in this context.


\subsection{Exploitation}
In BBO, pure exploitation corresponds to selecting $\b x$ to maximize the expected probability of success, $\mu_c(\boldsymbol{x}) = \Phi\br{\mu_f(\boldsymbol{x})\Big/\sqrt{1+\sigma_f(\boldsymbol{x})^2}}$. This differs from BO with continuous outputs, where pure exploitation corresponds to maximizing the posterior mean of $f\br {\b x}$, defined as $\mu_f\br{\b x}$ \citep{Snoek2012}. 

 In PBO, most authors define exploitation as selecting one duel member to maximize the mean of the objective function, $\mu_f\br{\b x}$ \citep{Brochu2010b, Nielsen2015a, Benavoli2020a}. Indeed, it is straightforward to show that $\boldsymbol{x}^\star=\underset{x\in \mathcal{X}}{\arg \max} \ \mu_f(\boldsymbol{x})$ is the most likely winner of any duel, regardless of the other duel member (since $P(c = 1 | [\boldsymbol{x}^\star, \boldsymbol{x}], \mathscr{D}) \geq \frac{1}{2}$ for all $\b x$). 
 Note that in PBO it makes little sense to choose both duel members according to the same exploitative strategy, as in this case $\b x = \b x'$, giving uninformative comparisons.

\subsection{Exploration}
\label{exploration}
Exploration is typically thought of as choosing values of $\b x$ for which uncertainty is maximal. In standard BO with noiseless observations, pure exploration corresponds to choosing $\b x$ to maximize the posterior variance over $f\br{\b x}$. In BBO we propose that it makes more sense to define uncertainty in the space of the stochastic binary observations, $c$. Here, there are two sources of uncertainty: epistemic uncertainty, which comes from our uncertainty about the objective function, $f\br{\b x}$, and aleatoric uncertainty, which remains even if we have complete knowledge of $f\br{\b x}$, and comes from the fact the $c$ is a stochastic variable. Further, with binary outputs, the level of aleatoric uncertainty is not constant but depends on the objective, $f\br{\b x}$.

For BBO, we argue that `exploration' should seek to maximize the epistemic uncertainty about the observed variable, $c$ \citep{Gonzalez2017a}. To motivate this choice, let us consider two cases. First consider the case where we have inferred that $f\br{\b x}\approx0$ to high precision, for a given $\b x$ (and thus, that $P\br{c=1|\b x}\approx 0.5$). Here, the aleatoric uncertainty about $c$ is near maximal while the epistemic uncertainty is close to zero. Thus, observing $c$ gives little new information about the objective, $f$. Second, consider the case where we are confident that $f\br{\b x}\gg 0$, despite being highly uncertain about its exact value. In this case, as $P\br{c=1}\approx 1$, we expect to gain little information about $f$ from individual observations of $c$. Despite the large uncertainty about $f$, the epistemic uncertainty about $c$ is small.

In the following, we present two alternative definitions of epistemic uncertainty, and show how they can be efficiently computed for use in GP classification. 

\subsubsection{Variance-based measure of uncertainty}
\label{variance_based}

First, we use variance as a measure of uncertainty \citep{Depeweg2017}. The total predictive variance, $\mathbb{V}(c|\mathscr{D},\boldsymbol{x})$ corresponds to the uncertainty about $c$, given $\boldsymbol{x}$.  In Bayesian learning, both aleatoric and epistemic uncertainties are modeled using the same probability measure, but they can be distinguished by applying the law of total variance: 
\begin{equation}
    \underbrace{\mathbb{V}(c|\mathscr{D}, \boldsymbol{x})}_{\textrm{total}}= \underbrace{\mathbb{V}_{p(f|\mathscr{D})}[\Phi(f(\boldsymbol{x}))]}_{\textrm{epistemic}} + \underbrace{\mathbb{E}_{p(f|\mathscr{D})}[\mathbb{V}(c|f, \boldsymbol{x})]}_{\textrm{aleatoric}}\label{variance_composition}.
\end{equation}
The first term on the right hand side of Eqn \ref{variance_composition}, corresponds to the posterior variance in the `success rate', $P\br{c=1| f\br{\b x}}=\Phi\br{f\br{\b x}}$. It comes from our epistemic uncertainty about the objective function, $f$. By definition:
\begin{equation}
\mathbb{V}\left[\Phi(f(\boldsymbol{x}))\right] =\int\left(\Phi(f(\boldsymbol{x}))-\mu_c(\boldsymbol{x})\right)^{2} p(f \mid \mathscr{D},\boldsymbol{x}) df .
\end{equation}
While previous work \citep{Gonzalez2017a} approximated this epistemic uncertainty using Monte-Carlo methods, we found that it could be expressed analytically as follows (see supplementary section \ref{variance_composition}):
\begin{equation}
\begin{aligned}
\mathbb{V}\left[\Phi\left(f(\boldsymbol{x})\right)\right] &= \mu_c\br{\b x}\br{1 - \mu_c\br{\b x}}  \\
&- 2 T\left(\frac{\mu_f(\boldsymbol{x})}{\sqrt{1+\sigma_f^2(\boldsymbol{x})}}, \frac{1}{\sqrt{1+2\sigma_f^2(\boldsymbol{x})}}\right)   ,
\label{eq:epistemic_uncertainty_analytic}
\end{aligned}
\end{equation}
where we recall  that  $\mu_c\br{\b x}\equiv \Phi\br{\frac{\mu_f\br{\b x}}{\sqrt{1+\sigma_f^2\br{\b x}}}}$, and  $T$ denotes the Owen's T-function, that can be cheaply evaluated numerically \citep{Young1974, Thomas1986}. The computation of the gradient with respect to $\boldsymbol{x}$ is straightforward (see supplementary section \ref{app:GradientVariance}).

When the objective function, $f$, is known to high precision,  the epistemic uncertainty goes to zero, and we are left only with the second term of Eqn \ref{variance_composition}, the aleatoric uncertainty, which is defined in general as:
$$\mathbb{E}_{p(f|\mathscr{D})}[\mathbb{V}(c|f, \boldsymbol{x})] = \mathbb{E}_{p(f|\mathscr{D})}[\Phi(f(\boldsymbol{x}))(1-\Phi(f(\boldsymbol{x})))].$$ The aleatoric uncertainty is largest when $\Phi\br{f\br{\b x}}$ has a high probability of being close to 0.5, and smallest when it has a high probability of being close to 0 or 1. We can then combine Eqns \ref{variance_composition} and \ref{eq:epistemic_uncertainty_analytic} to arrive at the following analytic expression for the aleatoric uncertainty:
\begin{equation}
\mathbb{E}_f[\mathbb{V}(c|f,\boldsymbol{x})]  = 2 T\left(\frac{\mu_f(\boldsymbol{x})}{\sqrt{1+\sigma_f^2(\boldsymbol{x})}}, \frac{1}{\sqrt{1+2\sigma_f^2(\boldsymbol{x})}}\right). 
\end{equation}

\subsubsection{Entropy-based measure of uncertainty}
\label{entropy_based}
Alternatively, we can quantify uncertainty using entropy. The total uncertainty about the observed output is $H(c|\mathscr{D}, \boldsymbol{x})$. Aleatoric uncertainty, on the other hand, is the uncertainty about the output when the underlying function is known, averaged over possible latent function values, that is $\mathbb{E}_{p(f|\mathscr{D})}[H(c|f,\boldsymbol{x}) ]$. The difference between these two quantities is defined as the epistemic uncertainty. However, we note that $H(c|\mathscr{D}, \boldsymbol{x}) - \mathbb{E}_{p(f|\mathscr{D})}[H(c|f,\boldsymbol{x}) ]$ is also the mutual information  between the output and the latent function: $I(c,f|\boldsymbol{x},\mathscr{D})$. 
\citet{Depeweg2017}  thus proposed the following decomposition:  
\begin{equation}
\label{entropy_decomposition}
\underbrace{H(c|\mathscr{D}, \boldsymbol{x})}_{\scriptstyle \text{total}}  = \underbrace{I(c, f|\mathscr{D}, \boldsymbol{x})}_{\scriptstyle \text{epistemic}} + \underbrace{\mathbb{E}_{p(f|\mathscr{D})}[H(c|f,\boldsymbol{x}) ]}_{\scriptstyle \text{aleatoric}}
\end{equation}
When making an observation at a point $\boldsymbol{x}$, the mutual information $I(c, f|\mathscr{D}, \boldsymbol{x})$ is the expected decrease in entropy of $f$ that is due to observing $c$ :
\begin{equation}
I(c, f|\mathscr{D}, \boldsymbol{x}) = H(f|\mathscr{D}, \boldsymbol{x}) - \mathbb{E}_{P(c|\mathscr{D}, \boldsymbol{x})}[H(f|c,\boldsymbol{x}) ] 
\end{equation}
As a consequence, selecting the point maximizing epistemic uncertainty corresponds to selecting the point maximizing the expected decrease in latent function entropy. 
In GP classification and preference learning,  $I(c,f|\boldsymbol{x},\mathscr{D})$ can be approximated and maximized at low cost \citep{Houlsby2011}. 

\begin{figure}[tb]
\centering
\includegraphics[width = \columnwidth]{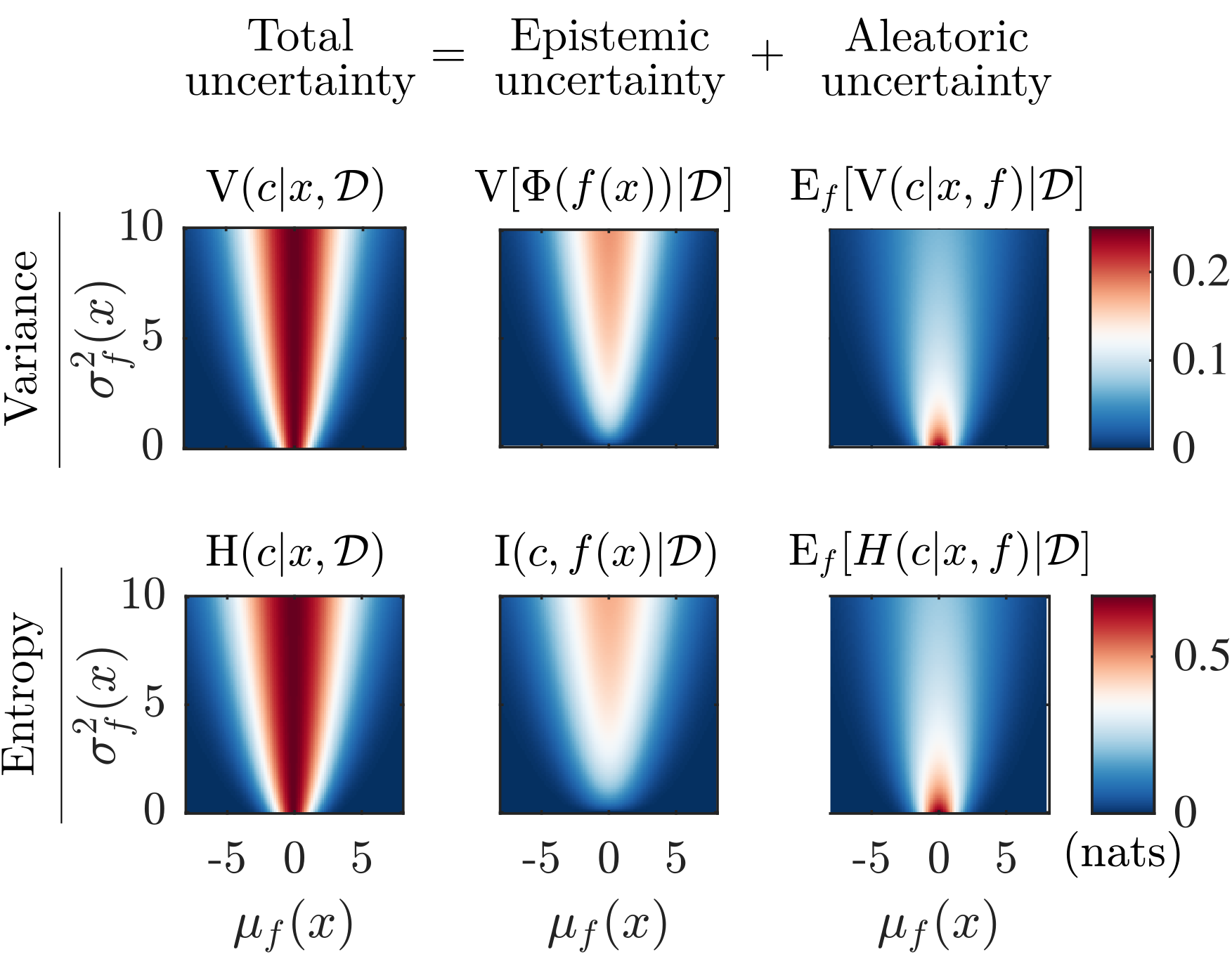} 
\caption[Uncertainty decomposition.]{Uncertainty decomposition in GP classification.  Horizontal and vertical axes correspond to the posterior mean $\mu_f(\boldsymbol{x})$ and variance $\sigma_f^2(\boldsymbol{x})$ of the latent GP, respectively. The first (resp.~second) row corresponds to uncertainty decomposition using variance (resp.~entropy).  (First column) Total uncertainty, quantified as: (above) the variance of observations, $c$; (below) the total entropy of $c$. (Second column)  Epistemic uncertainty, quantified as: (above) the posterior variance of $\Phi(f(\boldsymbol{x}))$; (below) the mutual information between $c$ and $f$. (Third column): Aleatoric uncertainty, quantified as: (above) the variance of $c$, given $f$; (ii) or the conditional entropy of $c$ given $f$).  We observe a high similarity between the both  uncertainty measures (compare upper and lower rows). Both epistemic and aleatoric uncertainty increase as $\mu_f$ approaches 0. However, aleatoric uncertainty decreases with $\sigma^2_f$ while epistemic uncertainty increases. Epistemic uncertainty is related to how informative a function evaluation is expected to be. In GP classification, contrary to standard GP regression,  epistemic uncertainty depends on the posterior mean of $f$.}
\label{fig:Uncertainties}
\end{figure}

The two alternative decompositions of uncertainty, plotted in figure \ref{fig:Uncertainties}, show qualitatively similar behaviour. We herein use the variance-based measure of uncertainty, since it could be computed exactly. However, due to the similarity between both measures (Fig \ref{fig:Uncertainties}), qualitatively similar results could be obtained with the entropy-based measure defined above.


\section{Acquisition functions }
Armed with a clear definition of exploitation and exploration, we can now propose new acquisition functions  for Bayesian optimization with binary outputs, that appropriately trade-off these two objectives.

\subsection{Upper Credible Bound in the space of observations}
A popular acquisition for standard BO with continuous noiseless outputs, is the upper credible bound (UCB):
\begin{equation}
    \alpha_{\mathrm{UCB}}\br{\b x} = \mathbb{E}[f\br{\b x}] + \beta \sqrt{\mathbb{V}\sqbr{f(\b{x})}},
\end{equation}
where the first term is the posterior mean of $f$, encouraging exploitation, and the second term is the posterior standard deviation of $f$, encouraging exploration. A constant, $\beta$, controls the trade-off between both terms. 

When used for BBO, we observed that this acquisition rule can lead to sub-optimal under-exploratory behavior. For example, it tends to select points with high latent mean, despite the fact that the output is very likely to be $c=1$ (even with large posterior variance over $f$), and thus, observing $c$ typically provides little information about $f$ (see section \ref{exploration}). 

This problem could be overcome by defining exploitation and exploration in the space of the observed variable, $c$, as described in the previous section, resulting in the following acquisition function:
\begin{equation}
\alpha_{\mathrm{UCB}_\Phi} (\boldsymbol{x})= \mathbb{E}[\Phi(f(\boldsymbol{x})] + \beta \sqrt{\mathbb{V}[\Phi(f(\boldsymbol{x}))]}
\label{eq:UCB}
\end{equation}
where the second term is the epistemic uncertainty, defined in section \ref{variance_based}. From sections \ref{exploitation} and \ref{variance_based}, we can write an analytic expression for this acquisition function and its gradient (supplementary \ref{app:GradientVariance}).  

\subsection{Maximally Uncertain Challenge for PBO}
 We propose a new acquisition rule for PBO, called the `Maximally Uncertain Challenge' (MUC), in which the first member of the duel, $\b x_1$ (the `champion'), exploits, while the second member, $\b x_2$ (the `challenger'), explores. As described in section \ref{exploitation}, in PBO a purely exploitative strategy for the first member of the duel corresponds to maximizing the mean of $f\br{\b x}$: 
\begin{equation}
\b{x}_1 \leftarrow \underset{\b{x} \in \mathcal{X}}{\arg \max }  \ \mu_f (\boldsymbol{x}).
\end{equation}
The next duel member is then chosen  according to a purely explorative strategy, to maximize the epistemic uncertainty, as defined in section 3.2.1:
\begin{equation}
\b{x}_2 \leftarrow \underset{\boldsymbol{x} \in \mathcal{X} }{\arg \max} \ \mathbb{V}[\Phi\br{f\br{\boldsymbol{x}_1}-f\br{\boldsymbol{x}}}].
\end{equation}

\subsection{Generalization to batches}
\label{sec:batch_PBO}
In batch preferential optimization, the user is asked to compare $m$ configurations, where $m>2$ \citep{Sui2017a, Siivola2020}. Performing multiple comparisons allows one to gather more information at each query, resulting in faster optimization. 

Several types of feedbacks can be gathered. Here, we consider the case where we collect comparisons between each pair of variables. In this case, the output of each comparison $(i,j)$ can be treated as a binary variable, $c_{ij} \in \{0,1\}$, sampled from a distribution:
\begin{equation}
    P\br{c_{ij}|f\br{\b x_i}, f\br{\b x_j}} = \Phi\br{f\br{\b x_i}- f\br{\b x_j}}
\end{equation}
Since, conditioned on $f$, observations are independent:
\begin{equation}
    P\br{\b c|f\br{\b x_1}, \dots, f\br{\b x_m}} = \prod_{i=1}^{m-1} \prod_{j>i}P\br{c_{ij}|f\br{\b x_i}, f\br{\b x_j}},
\end{equation}
where $\b c$ is a vector containing the output of all pairwise comparisons. In this case,  the problem reduces to a standard preference learning problem with $m\br{m-1}/2$ pairwise comparisons, and can be treated using the same inference methods as standard GP preference learning (section 2.2). 

The Maximally Uncertainty Challenge algorithm, that we developed for PBO, can be directly extended to deal with batch learning. To do this, we choose the first member of a batch (the `champion') according to a purely exploitative strategy, as follows:
\begin{equation}
\boldsymbol{x}_1 \leftarrow \underset{\boldsymbol{x} \in \mathcal{X}}{\arg \max }  \ \mu_f (\boldsymbol{x})
\end{equation}
Next, the other members of the batch (the `challengers') are chosen jointly, to maximize the sum of the epistemic uncertainty over all pairwise comparisons:
\begin{equation}
(\boldsymbol{x}_2,\cdots, \boldsymbol{x}_m) \leftarrow \underset{\boldsymbol{x}_2,\cdots, \boldsymbol{x}_m \in \mathcal{X} }{\arg \max} \ \sum_{i=1}^{i=m}\sum_{j > i}\mathbb{V}[\Phi(g(\boldsymbol{x}_i,\boldsymbol{x}_j))]
\end{equation}
Thus, we select challengers for which we are maximally uncertain about the outcome of the duel with $\boldsymbol{x}_1$ as well as the duels between challengers. An alternative approach would be to select $(\boldsymbol{x}_2, \cdots, \boldsymbol{x}_m)$ by maximizing $I(\boldsymbol{c}, f | \boldsymbol{x}_2, \cdots, \boldsymbol{x}_m)$. However, the computation of this term would be prohibitively expensive as batch size increases.  

\section{Results}
We evaluated the various acquisition functions on a set of 34 functions from a widely used virtual library for optimization experiments \citep{Surjanovic}. The functions in this library exhibit a diversity of behaviors that occur in real-life optimization problems. We compared the acquisition rules we propose to state-of-the-art heuristics. We used the stratified analysis method by \citet{Dewancker2016} to perform statistical analysis of acquisition functions performance on the benchmarks.

For each benchmark, we performed pairwise comparisons between acquisition functions using the Mann-Whitney U test at $\alpha = 5\times 10^{-4}$ significance on the best value found at the end of the optimization sequence. This determines a partial ranking based on the number of wins. Ties are broken by running the same procedure, but based on the Area Under Curve, which is related to the speed at which the algorithm reaches the optimum. This generates a new partial ranking, based on which a Borda score \citep{Dwork2001} is attributed to each acquisition function (the Borda score of any given candidate is the number of candidates with a lower rank). Rankings from different benchmarks are aggregated by summing the Borda scores to establish a global ranking. This can be seen as a weighted vote from each benchmark function. An advantage of this method is that it allows aggregating the results over a large set of benchmark functions.

 For each objective function, we inferred the hyperparameters for three different kernels (squared exponential, Matérn 3/2 and Matérn 5/2) using maximum likelihood estimation with 1000 randomly chosen samples. We then determined for each function the kernel that best described the function by measuring the root-mean-squared error on 3000 points. The benchmark functions are listed in supplementary \ref{app:benchmarks}. In all cases, we used the Expectation Propagation algorithm to approximate the posterior over the objective, $p\br{f|\mathscr{D}}$.

 Acquisition functions based on Thompson sampling require to sample from the GP posterior. We adapted previously published approximate sampling methods \citep{Solin2020a, Wilson2020a} to GP classification and preference learning (details in supplementary \ref{approximate_sampling}).
 
\subsection{Binary Bayesian optimization }
\begin{figure*}[hbt]
\centering
\includegraphics{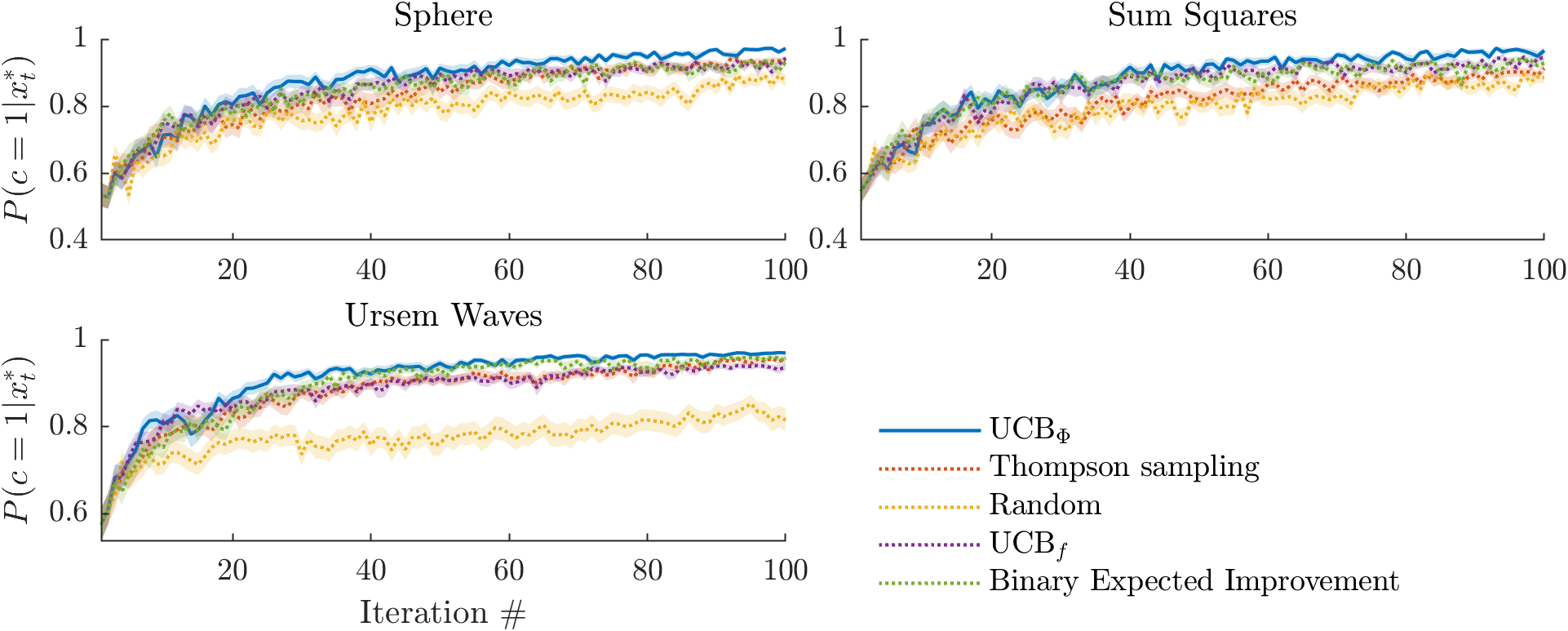}
\caption[BBO trajectories]{Evolution of the value of the inferred maximum $g(\boldsymbol{x}^\star)$ in three experiments of binary Bayesian optimization, repeated 60 times for 100 iterations. The ranking between acquisition functions varies depending on the objective function, however, statistical analysis (table \ref{tab:BBO_results}) shows that UCB$_\Phi$ and EI$_\Phi$ find better estimates of the maximum more rapidly.}
\label{fig:BBO_trajectories}
\end{figure*}

To avoid saturation effects when transforming the benchmark functions through the non-linearity $\Phi$, we scaled the test functions so that their mean is 0 and variance 1. For each test function, the experiment was repeated 60 times for 100 iterations. The initial two samples were selected at random. For $UCB_\Phi$, we chose a constant $\beta = \Phi^{-1}(0.99)$, whereas for $UCB_f$, following \citet{Tesch2013}, we used $\beta =1$.
 The results are presented in table \ref{tab:BBO_results}. More detailed results showing pairwise comparisons between acquisition functions are presented in the supplementary \ref{fig:BBO_benchmarks_matrices}. The evolution of the value of the inferred maxima is plotted in figure \ref{fig:BBO_trajectories} for three examples.

\begin{table}[!htb]\centering
\caption[Comparison of acquisition rules on benchmarks in BBO]{Comparison of acquisition functions on benchmarks in binary Bayesian optimization. }
\begin{tabular*}{\columnwidth}{@{\extracolsep{\fill}} ccc}
Acquisition rule & Rank & Borda score \\ 
\toprule
UCB$_\Phi$                                    & 1 & 77 \\ 
UCB$_f$ \citep{Tesch2013} & 2 & 45 \\ 
Thompson sampling                             & 3 & 41 \\ 
Binary EI \citep{Tesch2013}                     & 4 & 40 \\ 
Random                                        & 5 & 12 \\ 
\bottomrule
\end{tabular*}
\label{tab:BBO_results}
\end{table}

We found that, according to the Borda score, the UCB$_\Phi$ that we introduced outperforms all the other acquisition functions. In particular, in agreement with the results obtained by \citet{Tesch2013} for expected improvement (EI), UCB defined in the latent space (UCB$_f$) does not perform well compared to its counterpart defined in the probability space (UCB$_\Phi$). However, by contrast with the results obtained by \citet{Tesch2013} on 1D functions, the binary version of expected improvement (binary EI) does not perform well compare to $UCB_f$.
We did not systematically investigate the influence of the value of $\beta$ on the performance of UCB$_\Phi$. However, preliminary work suggests that performance is robust to changes in $\beta$.

\subsection{Preferential Bayesian optimization}
\label{subsec:results_PBO}

We repeated each experiment 40 times, for 80 iterations. The initial 5 samples were selected uniformly at random.
The results are presented in table \ref{tab:PBO_results}.
Again, more detailed results showing pairwise comparisons between acquisition functions as well as examples of regret curves  are presented in the supplementary \ref{supplementary}. The Maximally Uncertain Challenge acquisition rule outperforms all the other rules, with the exception of the dueling UCB rule which performs similarly. Acquisition rules based on Thompson sampling (TS, \citet{Thompson1933}) have limited performance. Expected Improvement performs only slightly better than random. This is likely because of the frequent occurrence of pathological behavior where the function samples identical duel members \citep{Nielsen2015a}.

\begin{table}[!htb]
\centering
\caption[Comparison of acquisition rules on benchmarks for PBO]{Comparison of acquisition functions on benchmarks in  preferential Bayesian optimization.}\label{tab:PBO_results}
\begin{tabular*}{\columnwidth}{@{\extracolsep{\fill}} ccc}
Acquisition rule & Rank & Borda \\ 
\toprule
Maximally Uncertain Challenge                              & 1 & 141 \\ 
Dueling UCB \citep{Benavoli2020a}                           & 1 & 141 \\ 
Bivariate EI \citep{Nielsen2015a}         & 3 & 111 \\ 
Duel TS \citep{Gonzalez2017a} & 4 & 106 \\ 
Dueling TS \citep{Benavoli2020a}                      & 5 & 101 \\ 
KSS  \citep{Sui2017a}                         & 6 & 66 \\ 
EIIG \citep{Benavoli2020a}                                  & 7 & 32 \\ 
EI \citep{Brochu2010a}                   & 8 & 23 \\ 
Random                                                     & 9 & 21 \\ 
\bottomrule
\end{tabular*}
\end{table}

\subsection{Batch learning}
To compare the batch variant of MUC to KernelSelfSparring (KSS, \citep{Sui2017a}), we ran batch PBO with $m=3$ for 30 iterations and 30 repetitions. The results are presented on table \ref{tab:batch_PBO_results} and figure \ref{fig:batch_PBO_trajectories}. Again, more detailed comparisons between acquisition rules are presented in supplementary section \ref{fig:batch_PBO_benchmarks_matrices}. The generalization of MUC outperforms KSS. This is likely because KSS selects batch members independently from each other, causing redundancy (for example, a duel could contain identical challengers).

\begin{figure*}[htb]
\centering
\includegraphics{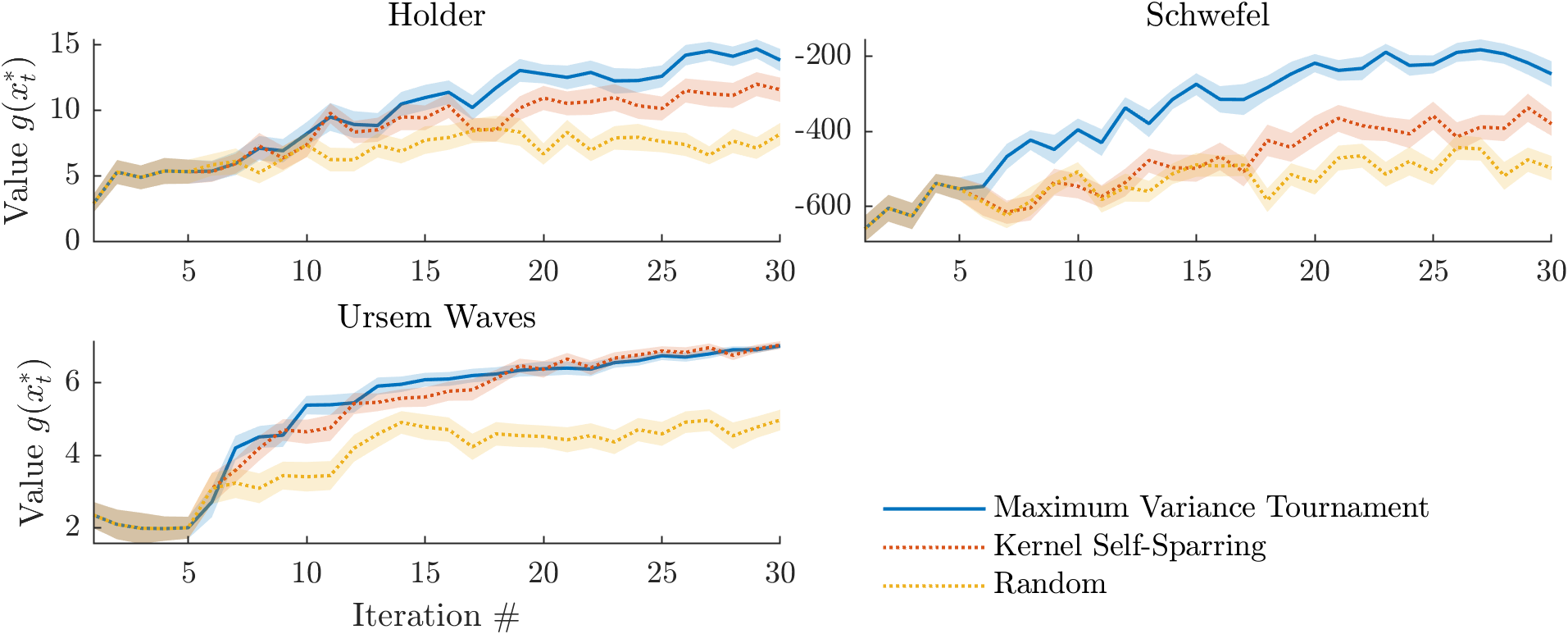}
\caption[Batch PBO trajectories]{Evolution of the value of the inferred maximum $g(\boldsymbol{x}^\star)$ in three experiments of batch preferential Bayesian optimization, repeated 30 times on 30 iterations. The ranking between acquisition functions varies depending on the objective function, however, statistical analysis (table \ref{tab:batch_PBO_results}) shows that over the whole functions set, the Dueling UCB and Maximally Uncertain Challenge find better estimates of the maximum more rapidly.}
\label{fig:batch_PBO_trajectories}
\end{figure*}

\begin{table}[!htp]\centering
\caption[Comparison of acquisition rules on benchmarks for batch PBO]{Comparison of acquisition functions on benchmarks for batch preferential optimization (batch size $m=3$).}
\begin{tabular*}{\columnwidth}{@{\extracolsep{\fill}} cccc}
Acquisition rule & Rank  & Borda score\\ 
\toprule
 MUC        & 1 & 32  \\ 
KSS \citep{Sui2017a} & 2 & 22 \\ 
Random                             & 3 & 0\\ 
\bottomrule
\end{tabular*}
\label{tab:batch_PBO_results}
\end{table}

\section{Discussion}
\subsection{Summary of contributions}
In this work, we made the following contributions : 
\begin{itemize}
\item We proposed an analytically tractable decomposition of uncertainty in GP classification into epistemic and aleatoric components.
\item By interpreting exploration as epistemic uncertainty sampling, we proposed new acquisition rules for Binary Bayesian optimization (UCB$_\Phi$) and
Preferential Bayesian optimization (the Maximally Uncertain Challenge), which outperform state-of-the-art heuristics.
\item By using a simple decomposition of batch preference learning into independent pairwise comparisons, we generalized the Maximally Uncertain Challenge to batch preferential optimization, where it shows superior performance compared to the state-of-the-art.
\end{itemize}
 
The definition of exploration as epistemic uncertainty sampling is very general. Beyond the binary and preferential setting, it would be interesting to apply it in the future to deal with Bayesian optimization with heteroscedastic noise \citep{Guzman2020}. 

We observed considerable variability in the performance of different acquisition functions depending on the objective function. This highlights the importance of using multiple benchmarks as well as statistical techniques for summarizing performance across various test functions. In contrast, previous work in this domain has often assessed the performance of different acquisition functions using only a few low-dimensional objective functions \citep{Tesch2013, Gonzalez2017a}.

  \subsection{Related work}
The decomposition of uncertainty into aleatoric and epistemic uncertainty is leveraged in many areas within machine learning \citep{Hullermeier2021}. \citet{Depeweg2017} first proposed the variance-based and entropy-based decompositions of uncertainty shown in Eqns~\ref{variance_composition} and \ref{entropy_decomposition}, in the context of  active learning with  Bayesian neural networks. Here we show how this uncertainty decomposition can be applied to GP classification, where it is possible to obtain analytical expressions for the variance-based measure (Eqn~\ref{eq:epistemic_uncertainty_analytic}), and an efficient approximation for the entropy-based measure \citep{Houlsby2011}.

\citet{Gonzalez2017a} proposed a PBO algorithm in which exploration is achieved by maximizing the epistemic uncertainty, $\mathbb{V}_{p(f|\mathscr{D})}[\Phi(f(\boldsymbol{x}))]$. However, in their work, this heuristic was not justified in terms of a decomposition of uncertainty into its aleatoric and epistemic components. Doing this allows us to: (i) draw a direct correspondence between variance-based epistemic uncertainty sampling (section \ref{variance_based}) and information-based measures of exploration (section \ref{entropy_based}) which have been used in the field of Bayesian active learning \citep{Houlsby2011}; and (ii) directly extend our principled definition of exploration and exploitation to develop new algorithms for BBO, PBO, and batch PBO. Finally, while \citet{Gonzalez2017a} used a Monte-Carlo approximation to estimate $\mathbb{V}_{p(f|\mathscr{D})}[\Phi(f(\boldsymbol{x}))]$, we derived an analytical expression, that could be easily computed (Eqn \ref{eq:epistemic_uncertainty_analytic}).



 
\citet{Tesch2013} noted that the performance of BBO is increased when the common expected improvement (EI) acquisition function is defined in the observation space, rather than the latent space. They hypothesized that this difference occurs because large values of the latent mean tend to be associated with high values of the latent variance, leading to over-exploitation of local maxima. However, from our experience with GP classification, there is no systematic association of high posterior latent mean with high posterior latent variance (see supplementary figure \ref{fig:BBO_proba_vs_latent}). We thus propose a different interpretation for their results: acquisition functions defined in the latent space tend to select points with high latent mean, despite the fact that, even even with large posterior latent variance, the output at these locations is very likely to be $c=1$,  and observations are expected to bring little information.

 In our proposed PBO acquisition rule the first duel member is chosen to exploit while the second member is chosen to explore. This rule performed significantly better than  Duel Thompson sampling \citep{Gonzalez2017a}, in which the first duel member is selected using TS, while the second member explores, by maximizing the epistemic uncertainty. Our rule also out-performed bivariate EI: a simple generalization of the popular EI acquisition function to the PBO setting \citep{Nielsen2015a}. 

 

 On the set of benchmark functions, dueling UCB, which selects the first duel member as the maximum of the posterior mean, and the second member using a UCB criterion defined in the latent space \citep{Benavoli2020a}, performed similarly to our proposed acquisition rule. In future work, it would be interesting to investigate whether modifying the dueling UCB rule to operate in the space of observations (rather than the latent space) could improve its performance, as we found to be the case for binary BO (Table \ref{tab:BBO_results}, Fig \ref{fig:BBO_trajectories}).  
 

Batch PBO was introduced in the GP bandits setting by \citet{Sui2017a}, who proposed the use of KernelSelfSparring, a variant of TS that involves sampling $m$ candidates from $p(\boldsymbol{x}^\star |\mathscr{D})$ (where $\b x^\star = \arg \max_x \mu_f\br{\b x}$) and making them compete. \citet{Siivola2020} independently proposed the same acquisition rule and showed that it outperformed the generalization of EI to batches. A drawback of this acquisition rule is that it selects duel members independently from each other, and thus, they are often highly correlated, resulting in under-exploratory behavior. Our method, where all but one member of the batch are chosen simultaneously, according to a single exploratory move, does not suffer from this limitation (see supplementary figure \ref{fig:batch_PBO}). In the future, it would be interesting to extend our method to alternative scenarios where: (i) we observe a single `winner' among the batch, or (ii) we obtain a ranking of the $k$ best alternatives \citep{Nguyen2020}.
 

 

The MATLAB code used in this work is available at \url{https://disclose_upon_release}.

\bibliography{Thesis_citations}
\bibliographystyle{icml2021}

\begin{appendices}
%

%

\onecolumn

\aistatstitle{Efficient exploration in binary and preferential Bayesian optimization: \\
Supplementary Materials}

\setcounter{equation}{0}
\setcounter{figure}{0}
\setcounter{table}{0}
\setcounter{section}{0}
\makeatletter
\renewcommand{\theequation}{S\arabic{equation}}
\renewcommand{\thefigure}{S\arabic{figure}}
\renewcommand{\thetable}{S\arabic{table}}
\renewcommand{\bibnumfmt}[1]{[S#1]}
\renewcommand{\citenumfont}[1]{S#1}

 \section{List of acquisition functions used in the PBO synthetic experiments}
 \label{app:acq_funs}
 \begin{table}[h]
 \centering
\scriptsize
\caption[List of acquisition rules for Preferential Bayesian Optimization.]{List of acquisition rules for Preferential Bayesian Optimization to which we compared our new method. In blue is the rule introduced in this paper.  Acquisition functions can be divided into classes. The first class corresponds to 'challenges', i.e. when the first duel member is the maximum of the posterior mean of the latent GP  $\boldsymbol{x}^{\star}$, sometimes restricted to training points (noted $\boldsymbol{x}^{\star\star}$). In the second class, the first duel member is selected via Thompson sampling. For both classes, the second duel member is selected using various heuristics that attempt to tradeoff exploration and exploitation.} 
\begin{tabular*}{\textwidth}{@{\extracolsep{\fill}}ccc}
\toprule
Name &First member &Second member \\ \toprule 
\begin{tabular}{@{}c@{}}Expected Improvement \\  \citep{Brochu2010a}\end{tabular}  & $\boldsymbol{x}^{\star\star} = \underset{x \in \mathcal{X}_t}{\arg \max} \ \mu_f(\boldsymbol{x})$ & $\underset{\boldsymbol{x} \in \mathcal{X}}{\arg \max } \ \mathbb{E}\left[\left[f(\boldsymbol{x}) - \mu_f(\boldsymbol{x}^{\star\star})\right]_+|\mathscr{D}_t\right]$\\
\begin{tabular}{@{}c@{}} Bivariate Expected Improvement \\ \citep{Nielsen2015a}\end{tabular} &$\boldsymbol{x}^{\star\star} = \underset{x \in \mathcal{X}_t}{\arg \max} \ \mu_f(\boldsymbol{x})$ & $\underset{\boldsymbol{x} \in \mathcal{X}}{\arg \max } \  \mathbb{E}\left[\left[f(\boldsymbol{x}) - f\left(\boldsymbol{x}^{\star\star}\right)\right]_+|\mathscr{D}_t\right]$ \\
{\color{blue}Maximally Uncertain Challenge (ours)} & $\boldsymbol{x}^{\star} = \underset{x \in \mathcal{X}}{\arg \max} \ \mu_f(\boldsymbol{x})$ & $\underset{\boldsymbol{x} \in \mathcal{X}}{\arg \max } \  \mathbb{V}\left[\Phi(g(\boldsymbol{x}^\star, \boldsymbol{x}))\right]$ \\
 \begin{tabular}{@{}c@{}} Dueling Thompson Sampling \\ \citep{Benavoli2020a} \end{tabular} &$\boldsymbol{x}^{\star} = \underset{x \in \mathcal{X}}{\arg \max} \ \mu_f(\boldsymbol{x})$ & Sample from $p\left(\boldsymbol{x}^\star|\mathscr{D}_t\right)$ \\ 
 \begin{tabular}{@{}c@{}} Dueling UCB \\ \citep{Benavoli2020a} \end{tabular} &$\boldsymbol{x}^{\star} = \underset{x \in \mathcal{X}}{\arg \max} \ \mu_f(\boldsymbol{x})$ & $\underset{\boldsymbol{x} \in \mathcal{X}}{\arg \max } \   [\mu_f(\boldsymbol{x}) + \beta \sigma_f(\boldsymbol{x})]$ \\ 
 \begin{tabular}{@{}c@{}} EI with Information Gain \\ \citep{Benavoli2020a} \end{tabular} &$\boldsymbol{x}^{\star} = \underset{x \in \mathcal{X}}{\arg \max} \ \mu_f(\boldsymbol{x})$ & $\underset{\boldsymbol{x}_2 \in \mathcal{X}}{\arg \max } \ k\log(\mu_c(\boldsymbol{x}_2,\boldsymbol{x}^\star))-I(c,g|\mathscr{D}_t, \boldsymbol{x}^\star, \boldsymbol{x}_2)$ \\ 
\begin{tabular}{@{}c@{}}Duel Thomspon Sampling \\ \citep{Gonzalez2017a} \end{tabular} & $\boldsymbol{x} \sim p\left(\boldsymbol{x}^\star|\mathscr{D}_t\right)$ & $\underset{\boldsymbol{x} \in \mathcal{X}}{\arg \max } \  \mathbb{V}\left[\Phi(g(\boldsymbol{x}^\star, \boldsymbol{x}))\right]$ \\
  \begin{tabular}{@{}c@{}} Kernel Self Sparring \\ \citep{Sui2017a}\end{tabular}  & $\boldsymbol{x} \sim p\left(\boldsymbol{x}^\star|\mathscr{D}_t\right)$ & $\boldsymbol{x} \sim p\left(\boldsymbol{x}^\star|\mathscr{D}_t\right)$ \\
\bottomrule
\end{tabular*}
\label{tab:pref_acquisition_funs}
\end{table}

\clearpage
\section{Variance of the predictive class distribution}
\label{Variance}
By definition, the variance of the predictive class distribution is: 
\begin{equation}
\label{eq:var_muc}
\begin{aligned}
\mathbb{V}\left[\Phi(f(\boldsymbol{x}))\right] &=\int\left(\Phi(f(\boldsymbol{x}))-\mathbb{E}\left[\Phi(f(\boldsymbol{x}))\right]\right)^{2} p(f \mid \mathscr{D}_t,\boldsymbol{x}) d f 
\end{aligned}
\end{equation}As a consequence :
\begin{equation}
\label{eq:var_muc_developed}
\begin{aligned}
\mathbb{V}\left[\Phi(f(\boldsymbol{x}))\right] &=\int \Phi(f(\boldsymbol{x}))^{2} p(f(\boldsymbol{x}) \mid \mathscr{D}_t,\boldsymbol{x}) d f-\mathbb{E}\left[\Phi(f(\boldsymbol{x}))\right]^{2} \\
&=\int \Phi\left(\sigma_f(\boldsymbol{x}) f(\boldsymbol{x}) + \mu_f(\boldsymbol{x})\right)^{2} \phi \left(f(\boldsymbol{x})\right) d f-\mathbb{E}\left[\Phi\left(f(\boldsymbol{x})\right)\right]^{2}, 
\end{aligned}
\end{equation}
where $\phi$ is the probability density function of $\mathcal{N}\left(0,1\right)$,  $\mu_f$ and $\sigma^2_f$ are the mean and variance of the Gaussian approximation to the posterior  $p\left(f|\mathscr{D}_t\right)$.
Making use of the following formula \citep{Patel1996}:
\begin{equation}
\begin{aligned}
\int_{-\infty}^{\infty} \Phi\left(a+b t\right)^{2} \phi(t) d t=\Phi\left(\frac{a}{\sqrt{1+b^{2}}}\right)-2 T\left(\frac{a}{\sqrt{1+b^{2}}}, \frac{1}{\sqrt{1+2 b^{2}}}\right),
\end{aligned}
\end{equation}

with the Owen's T function defined as: 
\begin{equation}
\begin{aligned}
T\left(h, a\right)=\frac{1}{2 \pi} \int_{0}^{a} \frac{e^{-\frac{1}{2} h^{2}\left(1+t^{2}\right)}}{1+t^{2}} d t,
\end{aligned}
\end{equation}

equation \ref{eq:var_muc} can be written as: 

\begin{equation}
\label{eq:var_muc_analytic}
\begin{aligned}
\mathbb{V}\left[\Phi\left(f(\boldsymbol{x})\right)\right] &= \mathbb{E}\left[\Phi\left(f(\boldsymbol{x})\right)\right] - 2 T\left(\frac{\mu_f(\boldsymbol{x})}{\sqrt{1+\sigma_f^2(\boldsymbol{x})}}, \frac{1}{\sqrt{1+2\sigma_f^2(\boldsymbol{x})}}\right) - \mathbb{E}\left[\Phi\left(f(\boldsymbol{x})\right)\right]^{2} 
\end{aligned}
\end{equation}

\section{Gradient of the variance of the predictive class distribution}
\label{app:GradientVariance}

The derivatives of the Owen's T function are: 
\begin{equation}
\frac{\partial T}{\partial h} =  -\frac{1}{2 \pi} \int_{0}^{a} h e^{-\frac{1}{2} h^{2}\left(1+x^{2}\right)}dx =  -\frac{1}{2\pi}e^{-\frac{h^2}{2}}\int_0^{ha}e^{-\frac{t^2}{2}}dt= -\frac{e^{-\frac{h^2}{2}}}{\sqrt{2\pi}}\left(\Phi(ha)-\frac{1}{2}\right),
\end{equation}  
\begin{equation}
\frac{\partial T}{\partial a} =  \frac{1}{2 \pi}  \frac{e^{-\frac{1}{2} h^{2}\left(1+a^{2}\right)}}{1+a^{2}}.
\end{equation} 
Moreover, we have :  
\begin{equation}
\mathbb{E}_f\left[\Phi\left(f(\boldsymbol{x})\right)\right] = \Phi\left(\frac{\mu_f(\boldsymbol{x})}{\sqrt{1+\sigma^2_f(\boldsymbol{x})}}\right).
\end{equation}
So computing the gradient of $\mathbb{V}_f[\Phi(f(\boldsymbol{x}))|\mathscr{D}_t]$ is straightforward.

\clearpage

\section{Sampling from GP classification models}
\label{approximate_sampling}

Acquisition rules based on Thompson sampling \citep{Thompson1933, Russo2018}: KernelSelfSparring \citep{Sui2017a}, Dueling Thompson Sampling \citep{Benavoli2020a}, and Duel Thompson Sampling \citep{Gonzalez2017a}, rely on samples from the posterior distribution over the maximum (see table \ref{tab:pref_acquisition_funs}): 
\begin{equation}
p\left(\boldsymbol{x}^\star \mid \mathscr{D}\right)=p\left(f\left(\boldsymbol{x}^\star\right)=\max _{\boldsymbol{x} \in \mathcal{X}} f(\boldsymbol{x}) \mid \mathscr{D}\right)
\end{equation}
\citet{Hernandez-Lobato2014} proposed the following sampling scheme: draw a sample from the posterior distribution $p\left(f | \mathscr{D}\right)$, then return the maximum of the sample. One could iteratively construct the sample $f$ while it is being optimized but, as noted by \citet{Hernandez-Lobato2014}, this would have a cost $\mathcal{O}\left(m^3\right)$, where $m$ is the number of evaluations of the function necessary to find the maximum. Although this is doable in practice,  \citet{Hernandez-Lobato2014}  suggested a more efficient procedure by sampling a finite-dimensional approximation to $f$, based on a finite-dimensional approximation to the kernel $k\left(\boldsymbol{x},\boldsymbol{x'}\right) \sim \phi(\boldsymbol{x})^\top \phi\left(\boldsymbol{x'}\right)$ \citep{Lazaro-Gredilla2010}. In GP classification and preference learning, this approximate sampling cannot be directly applied. In the following section, we will explain how to apply existing approximate sampling methods to the case of GP classification models.   

\subsection{Kernel approximation}
The sampling methods mentioned above consists in approximating a stationary kernel $k$ by means of the inner product of features $\phi$ such that: $k(\boldsymbol{x},\boldsymbol{x'}) \sim \phi(\boldsymbol{x})^\top \phi(\boldsymbol{x}')$.  
Recently, a method was proposed by \cite{Solin2020a}, which aims at making the approximation as good as possible for a given rank (see \citet{Riutort-Mayol2020} for details about the practical implementation). 
In this method, the kernel is approximated using a series expansion in terms of eigenfunctions of the Laplace operator on a rectangular domain $\Omega = [-L_1,L_1] \times \cdots \times [-L_d,L_d]$ (the search space are usually rectangular in Bayesian optimization).

 In preference learning, a specific difficulty arises. Indeed, the base kernel used to model the value function may be shift-invariant, the preference kernel, however, is not in general. 
This inexact hypothesis introduced in the sampling algorithm, leads to samples that are not consistent with the anti-symmetric property of a preference function, i.e. $f(\boldsymbol{x},\boldsymbol{x}') = -f(\boldsymbol{x}',\boldsymbol{x})$,  (see  e.g.  figure 4.1 in  \citet{Gonzalez2017a} where this inexact stationarity hypothesis is introduced). 

However, assume that we have a finite dimensional approximation to the base kernel $k(\boldsymbol{x},\boldsymbol{x}') \sim \phi(\boldsymbol{x})^\top \phi(\boldsymbol{x}')$, it is easy to see that we can approximate the preference kernel by $k_{\text{pref}}((\boldsymbol{x}_i,\boldsymbol{x}_j),(\boldsymbol{x}_k, \boldsymbol{x}_l)) \sim \phi_{\text{pref}}(\boldsymbol{x}_i, \boldsymbol{x}_j)^\top \phi_{\text{pref}}(\boldsymbol{x}_k, \boldsymbol{x}_l) $, with:
\begin{equation}
\label{eq:preference_features}
\phi_{\text{pref}}(\boldsymbol{x}_i, \boldsymbol{x}_j)= \phi(\boldsymbol{x}_i) - \phi(\boldsymbol{x}_j)
\end{equation} 
By construction, the corresponding sample is anti-symmetric.

\subsection{Weight-space approximation with non-Gaussian likelihoods}

The most widely used method for approximate sampling from GP in Bayesian optimization is the weight-space approximation. Assume that we have a finite-dimensional approximation to the kernel $k\left(\boldsymbol{x},\boldsymbol{x'}\right) \sim \phi(\boldsymbol{x})^\top \phi\left(\boldsymbol{x'}\right)$. The features $\phi(\boldsymbol{x})$ can be used to approximate the Gaussian process posterior with a Bayesian linear model: $f(\boldsymbol{x})\sim \phi(\boldsymbol{x})^\top \omega$, where \citep{Lazaro-Gredilla2010}:

\begin{equation}
\omega \sim \mathcal{N}\left(\left(\Phi^\top \Phi + \sigma^2I\right)^{-1}\Phi^\top \boldsymbol{y}, \left(\Phi^\top \Phi + \sigma^2I\right)^{-1}\sigma^2\right) 
\label{eq:sampling_w}
\end{equation} In the case of non-Gaussian likelihoods, , naively replacing $\boldsymbol{y}$ in \ref{eq:sampling_w} by the latent values inferred by the Laplace approximation or Expectation Propagation would not take into account the correlated heteroscedastic noise on the latent function values at training points. 
To the extent of our knowledge, the process of weight-space approximate sampling has not been rigorously introduced for latent Gaussian process models. Here, we suggest to use a sampling process in two steps. First, samples $\boldsymbol{y}$ are drawn from the posterior distribution over the latent variables at training points: $\mathcal{N}(\boldsymbol{\mu}, \boldsymbol{\Sigma})$, then $\boldsymbol{\omega}$ is sampled from $\boldsymbol{\omega} \sim \mathcal{N}\left(\left(\Phi^\top \Phi + \sigma^2I\right)^{-1}\Phi^\top \boldsymbol{y}, \left(\Phi^\top \Phi + \sigma^2I\right)^{-1}\sigma^2\right) $, where $\sigma$ is a small constant used for regularization. 

To see why this sampling scheme is correct, note that:
$p(f(\boldsymbol{x})|\mathscr{D}) = \int p(f(\boldsymbol{x})|X, \boldsymbol{y})p(\boldsymbol{y}|\mathscr{D})d\boldsymbol{y}$. So given latent values sampled from $p(\boldsymbol{y}|\mathscr{D})$, an approximate sample $\widetilde{f}$ can be drawn from $p(f(\boldsymbol{x})|X, \boldsymbol{y})$ using the method of \citet{Hernandez-Lobato2014}. 

 However, the degeneracy, i.e., low-rankness of the GP approximation, causes the estimate to grow over-confident when the number of observed points exceeds the degrees of freedom of the approximation. This results in ill-behaved approximations, and, in particular, underestimated variance, in regions far away from the data points. This phenomenon is known as variance starvation \citep{Wang2018a, Mutny2018, Calandriello2019a}.

\subsection{Decoupled-bases approximate sampling}

Recently \citet{Wilson2020a} proposed an efficient way to sample from Gaussian process posteriors that avoids variance starvation. The original sampling method was devised for exact GP with Gaussian noise and sparse Gaussian processes, where the GP is computed based on a set of inducing points that explain the data, however, it can easily be generalized to non-Gaussian likelihood with a latent function \citep{Wilson2020b}. 

Briefly, this method is based on a corollary of Matheron's rule. For a Gaussian process $f \sim \mathcal{G} \mathcal{P}\left(0, k\right)$, the latent process conditioned on latent values  $(X, \boldsymbol{y})$  admits, in distribution, the representation: 
\begin{equation}
\begin{aligned}
\underbrace{\left(f \mid \boldsymbol{y}\right)\left(\cdot\right)}_{\scriptstyle \text {posterior}} \stackrel{\mathrm{d}}{=} \underset{\scriptstyle \text {prior}}{\underbrace{f\left(\cdot\right)}}+\underset{\scriptstyle \text {update}}{\underbrace{k\left(\cdot, \boldsymbol{x}\right) \mathbf{K}^{-1}\left(\boldsymbol{y}-f(X)\right)}}
\end{aligned}
\end{equation}

This corollary defines an approximation to the Gaussian process conditioned on $(X, \boldsymbol{y})$, where the stationary prior is approximated with a Bayesian linear model (weight-space prior), and the approximate posterior is obtained by adding an exact update (function-space update): 

\begin{equation}
\label{eq:Decoupled_bases}
\underbrace{\left(f \mid \boldsymbol{y}\right)\left(\cdot\right)}_{\scriptstyle \text{posterior}} \stackrel{\mathrm{d}}{\approx} \underbrace{\sum_{i=1}^{\ell} \omega_{i} \phi_{i}\left(\cdot\right)}_{_{\scriptstyle \text{weight-space prior}}}+\underbrace{\sum_{j=1}^{m} v_{j} k\left(\cdot, \boldsymbol{x}_{j}\right)}_{\scriptstyle \text{function space update}},
\end{equation}

 where $\boldsymbol{v}=\mathbf{K}^{-1}\left(\boldsymbol{y}-\boldsymbol{\Phi} \boldsymbol{\omega}\right)$, and $\boldsymbol{\omega}$ is sampled from $\mathcal{N}(\boldsymbol{0}, I)$. This method is termed decoupled-bases decomposition of the GP.
 
To sample from the posterior latent function, we thus sample   $\boldsymbol{\omega}$  from $\mathcal{N}(\boldsymbol{0}, I)$ and compute the corresponding weight-space prior, then sample $\boldsymbol{y}$ from $\mathcal{N}\left(\mu_f\left(X\right), \Sigma_f\left(X,X\right)\right)$ and compute the corresponding function-space update.
 
 \clearpage
 \section{Supplementary results}
\label{supplementary}
\begin{figure}[h]
\centering
\includegraphics{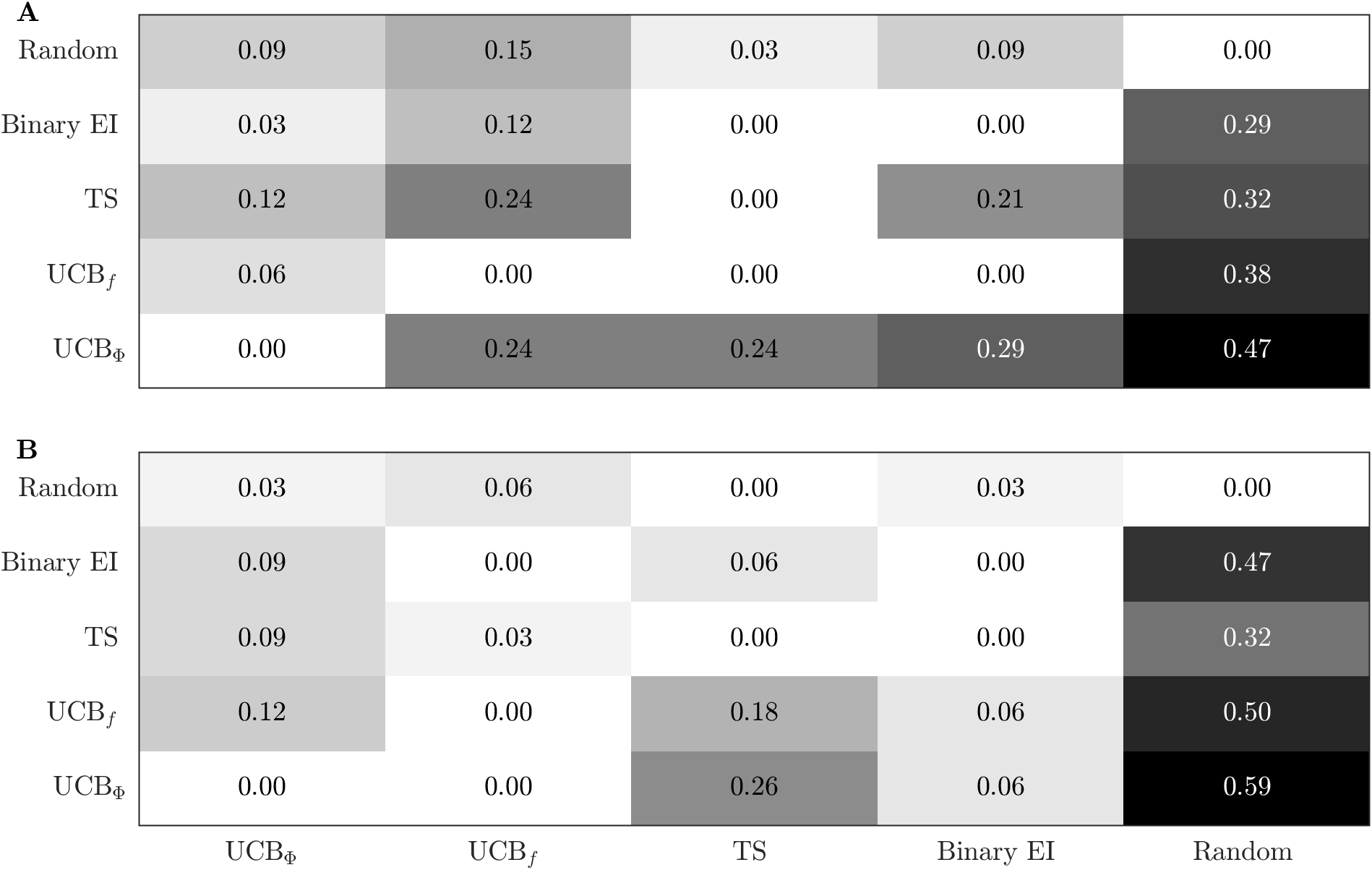}
\caption[Bayesian optimization]{Detailed results of performance comparison between acquisition functions in Binary Bayesian optimization. Each entry (i,j) corresponds to the fraction of benchmarks functions for which i beats j according to the Mann-Whitney U test at $\alpha = 5\times 10^{-4}$ significance based either on the best value found (\textbf{A}) or the Area Under the Curve (\textbf{B}).}
\label{fig:BBO_benchmarks_matrices}
\end{figure}

\begin{figure}[h]
\centering
\includegraphics{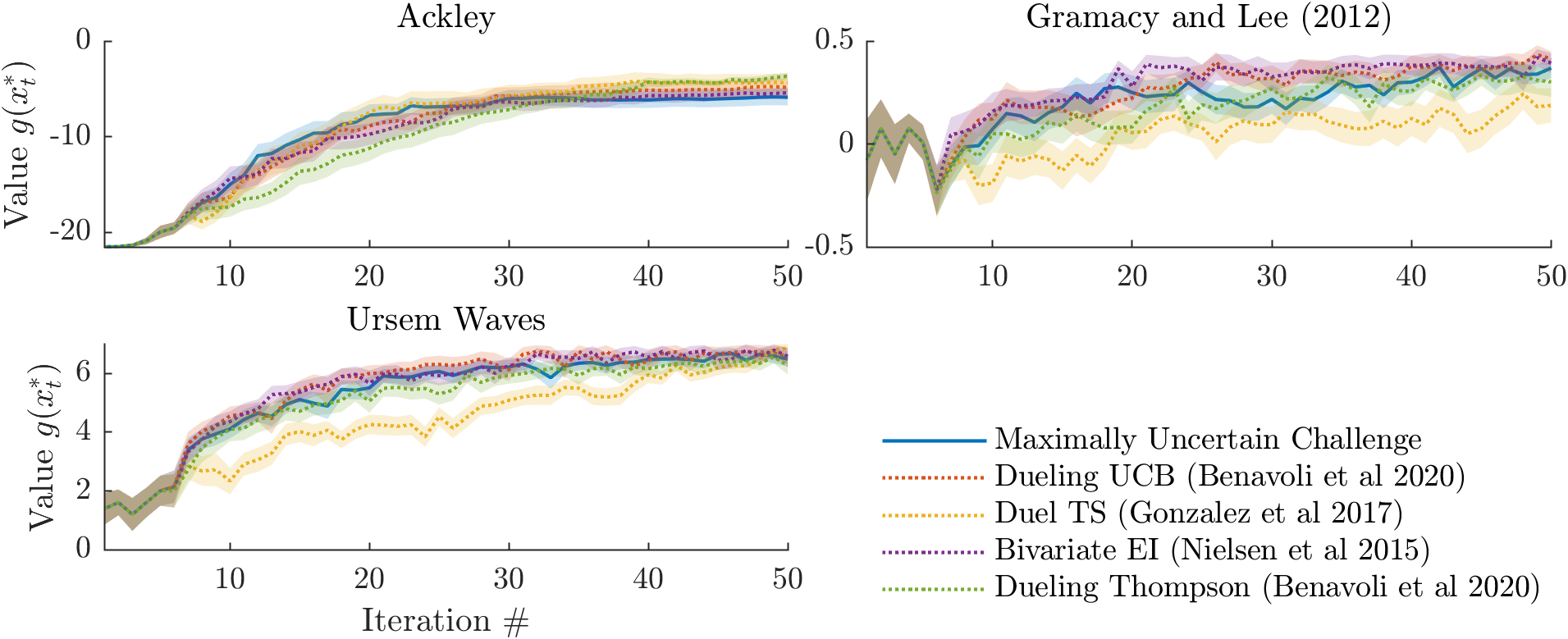}
\caption[PBO trajectories]{Evolution of the value of the inferred maximum $g(\boldsymbol{x}^\star)$ in three experiments of preferential Bayesian optimization, repeated 20 times on 50 iterations. The ranking between acquisition functions varies depending on the objective function, however, statistical analysis (table \ref{tab:PBO_results}) shows that over the whole functions set, the Dueling UCB and Maximally Uncertain Challenge find better estimates of the maximum more rapidly.}
\label{fig:PBO_trajectories}
\end{figure}

\begin{figure}[h]
\centering
\includegraphics{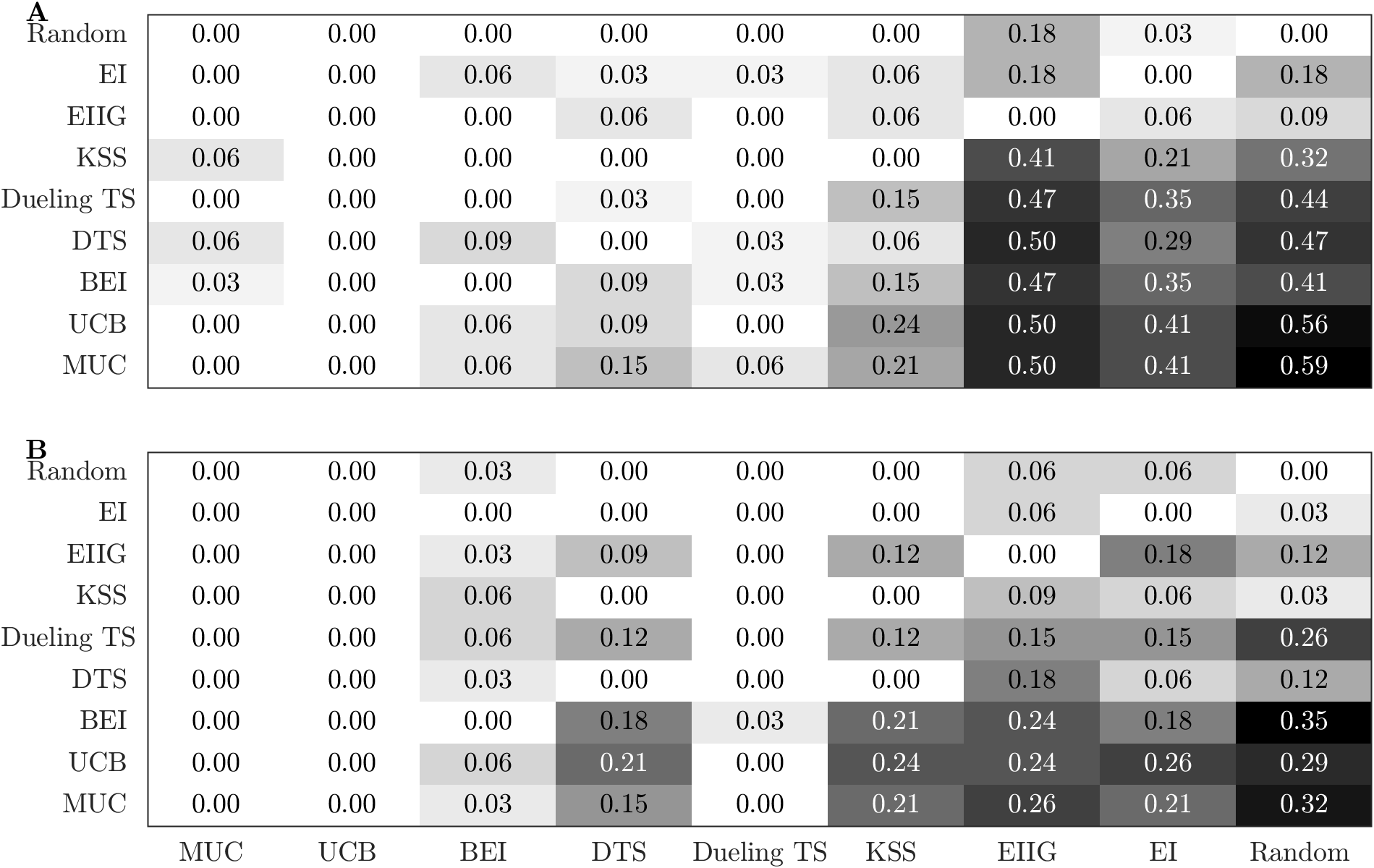}
\caption[Bayesian optimization]{Detailed results of performance comparison between acquisition functions in Preferential Bayesian optimization.Each entry (i,j) corresponds to the fraction of benchmarks functions for which i beats j according to the Mann-Whitney U test at $\alpha = 5\times 10^{-4}$ significance based either on the best value found (\textbf{A}) or the Area Under the Curve (\textbf{B}).}
\label{fig:PBO_benchmarks_matrices}
\end{figure}

\begin{figure}[h]
\centering
\includegraphics{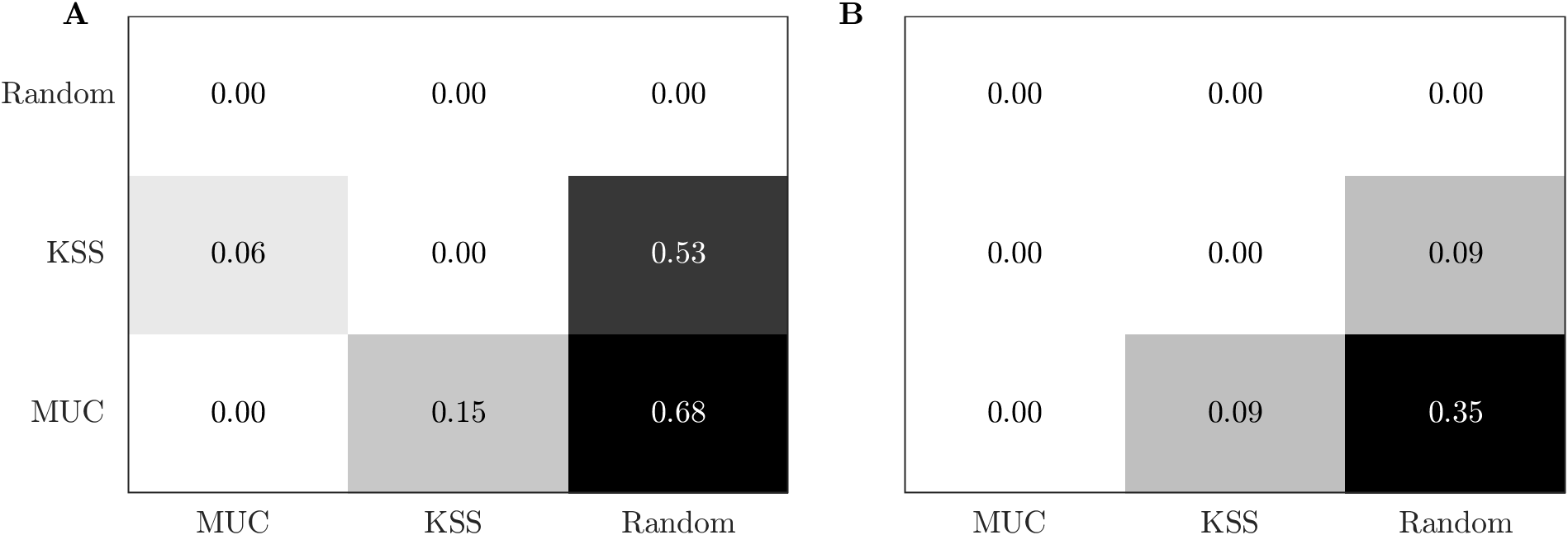}
\caption[Bayesian optimization]{Detailed results of performance comparison between acquisition functions in batch Preferential Bayesian optimization. Each entry (i,j) corresponds to the fraction of benchmarks functions for which i beats j according to the Mann-Whitney U test at $\alpha = 5\times 10^{-4}$ significance based either on the best value found (\textbf{A}) or the Area Under the Curve (\textbf{B}).}
\label{fig:batch_PBO_benchmarks_matrices}
\end{figure}

\clearpage

\section{Comparison of UCB in the observation and in the latent space}
\begin{figure}[h!]
\includegraphics[width=\linewidth]{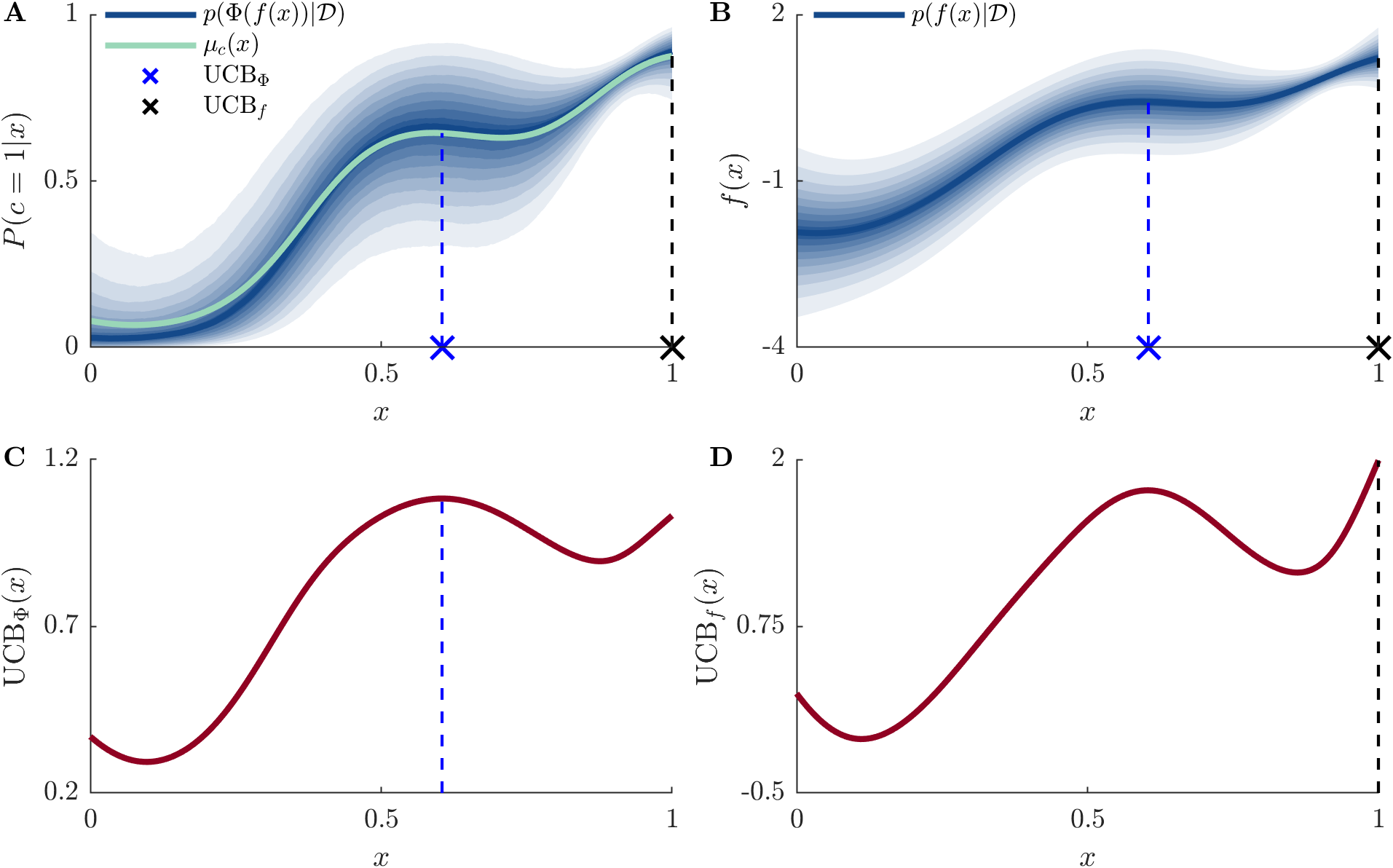}
\caption[Comparison between UCB in the latent and the probability space.]{Comparison between Upper Credible Bound in the latent space (UCB$_f$) and UCB in the observation space (UCB$_\Phi$). \textbf{A.} Posterior distribution $p(\Phi(f(x))|\mathscr{D})$ (in blue, shaded areas correspond to ventiles of the distribution), and predictive class distribution $\mu_c(\boldsymbol{x})$ (in turquoise). The choice of the next query according to UCB$_\Phi$ and UCB$_f$ is indicated by crosses. \textbf{B.} Posterior Gaussian process $p(f|\mathscr{D})$ (in blue). \textbf{C.} UCB acquisition function in the latent space. The selected point is indicated with a dashed line. \textbf{D.} UCB acquisition function in the latent space. The selected point is indicated with a dashed line.  
This example illustrates a pathological behavior of UCB$_f$, whereby the acquisition function tends to select points with high posterior latent mean, despite the fact that these points correspond to very little uncertainty about the output. This under-exploration problem does not appear with UCB$_\Phi$.} 
\label{fig:BBO_proba_vs_latent}
\end{figure}

\clearpage
\section{Benchmarks}
\label{app:benchmarks}
 
\centering
\begin{longtable}{lccc}
 Name & D & Kernel & Space \\ 
\toprule
\endhead
Ackley & 2 & Matérn 3/2 & $[-32.768,32.768]^2$ \\ 
Beale & 2 & SE-ARD & $[-4.5,4.5]^2$ \\ 
Bohachevsky & 2 & SE-ARD & $[-100,100]^2$ \\ 
Three-Hump Camel & 2 & Matérn 5/2 & $[-5,5]^2$ \\ 
Six-Hump Camel & 2 & SE-ARD & $[-3,3]\times [-2,2]$ \\ 
Colville & 4 & Matérn 5/2 & $[-10,10]^4$ \\ 
Cross-in-Tray & 2 & Matérn 5/2 & $[-10,10]^2$ \\ 
Dixon-Price & 2 & Matérn 5/2 & $[-5,5]^2$ \\ 
Drop-Wave & 2 & Matérn 3/2 & $[-5.12,5.12]^2$ \\ 
Eggholder & 2 & SE-ARD & $[-512,512]^2$ \\ 
Forrester et al (2008) & 1 & SE-ARD & $[0,1]$ \\ 
Goldstein-Price & 2 & SE-ARD & $[-2,2]^2$ \\ 
Griewank & 2 & SE-ARD & $[-600,600]^2$ \\ 
Gramacy and Lee (2012) & 1 & SE-ARD & $[0.5,2.5]$ \\ 
Hartmann 3-D & 3 & SE-ARD & $[0,1]^3$ \\ 
Hartmann 4D & 4 & SE-ARD & $[0,1]^4$ \\ 
Hartmann 6D & 6 & SE-ARD & $[0,1]^6$ \\ 
Holder & 2 & SE-ARD & $[-10,10]^2$ \\ 
Langer & 2 & Matérn 3/2 & $[0,10]^2$ \\ 
Levy & 2 & SE-ARD & $[-10,10]^2$ \\ 
Levy N.13 & 2 & Matérn 5/2 & $[-10,10]^2$ \\ 
Perm 0,d,$\beta$ & 2 & SE-ARD & $[-2,2]^2$ \\ 
Perm d,$\beta$ & 2 & SE-ARD & $[-2,2]^2$ \\ 
Powell & 4 & SE-ARD & $[-4,5]^4$ \\ 
Rosenbrock & 2 & SE-ARD & $[-2.048,2.048]^2$ \\ 
Rotated Hyper-Ellipsoid & 2 & Matérn 3/2 & $[-65.536,65.536]^2$ \\ 
Schaffer n4 & 2 & Matérn 3/2 & $[-100,100]^2$ \\ 
Schwefel & 2 & SE-ARD & $[-500,500]^2$ \\ 
Shekel & 4 & SE-ARD & $[0,10]^4$ \\ 
Schubert & 2 & Matérn 3/2 & $[0,10]^2$ \\ 
Sphere & 2 & SE-ARD & $[-5.12,5.12]^2$ \\ 
Sum Squares & 2 & SE-ARD & $[-10,10]^2$ \\ 
Trid & 2 & SE-ARD & $[-4,4]^2$ \\ 
Ursem Waves & 2 & SE-ARD & $[-1.2,1.2]\times [-0.9,1.2]$ \\ 
\bottomrule
\caption[Benchmarks]{Benchmark functions in Bayesian optimization experiments.}
\label{tab:benchmarks}
\end{longtable}
\clearpage

 \section{Comparison between KernelSelfSparring and MUC}
 \begin{figure}[h!]
\centering
\includegraphics{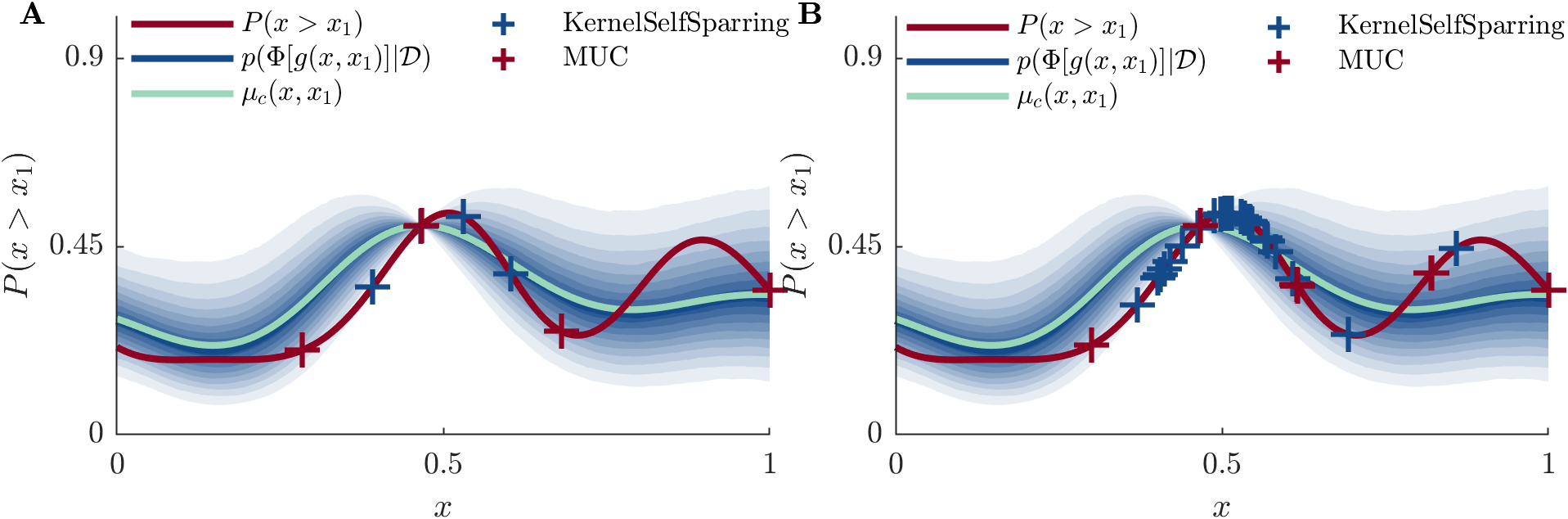}
\caption[Batch preferential Bayesian optimization]{Batch preferential Bayesian optimization. \textbf{A.} With batches of size 4. Note that the MUC acquisition rule is more explorative than KSS. \textbf{B.} With batches of size 25. For large batches, MUC tends to select the same challengers several times, which may be explained by the fact that it does not take correlation between duels into account (hence cannot prevent from selecting the same duels several times).}
\label{fig:batch_PBO}
\end{figure}
\vfill

\end{appendices}

\end{document}